\pdfoutput=1
\documentclass[sigconf,screen]{acmart}

\usepackage{balance}
\usepackage{graphicx}
\usepackage{colortbl}
\usepackage{arydshln}
\usepackage{times}
\usepackage{rotating}
\usepackage{makecell}
\usepackage{tabularx}
\usepackage{booktabs}
\usepackage{wrapfig}
\usepackage{tikz}
\usetikzlibrary{angles}
\usepackage{makecell}
\usepackage{tabu}
\usepackage{multirow}
\usepackage{hyperref}
\usepackage{framed}
\usepackage[framemethod=tikz]{mdframed}
\usetikzlibrary{shadows}
\usepackage{enumitem}
\usepackage{graphics}
\newmdenv[
tikzsetting= {fill=gray!10},
linewidth=1pt,
roundcorner=2pt,
shadow=false
]{myshadowbox}

\newenvironment{result}[2]
{\begin{myshadowbox}\textbf{\textit{\underline{Lesson#1:}}} #2}{
\end{myshadowbox}}

\hypersetup{
    linkcolor=blue,
    filecolor=magenta,      
    urlcolor=cyan,
}

\newcommand{\bi}{\begin{itemize}[leftmargin=0.4cm]}
\newcommand{\ei}{\end{itemize}}
\newcommand{\be}{\begin{enumerate}[leftmargin=0.4cm]}
\newcommand{\ee}{\end{enumerate}}

\usepackage{url}

\tikzstyle{thmbox} = [rectangle, rounded corners, draw=black, fill=gray!10]

\renewcommand{\textrightarrow}{$\rightarrow$}

\setcopyright{acmcopyright}
\acmPrice{15.00}
\acmDOI{10.1145/3278142.3278145}
\acmYear{2018}
\copyrightyear{2018}
\acmISBN{978-1-4503-6056-2/18/11}
\acmConference[SWAN '18]{Proceedings of the 4th ACM SIGSOFT International Workshop on Software Analytics}{November 5, 2018}{Lake Buena Vista, FL, USA}
\acmBooktitle{Proceedings of the 4th ACM SIGSOFT International Workshop on Software Analytics (SWAN '18), November 5, 2018, Lake Buena Vista, FL, USA}

\title{Is One
Hyperparameter Optimizer Enough?}
\author{Huy Tu}
\email{hqtu@ncsu.edu}
\author{Vivek Nair}
\email{vivekaxl@gmail.com}
\affiliation{ 
      \institution{North Carolina State University}
      \city{Raleigh} 
      \state{NC}
      \country{USA}
      \postcode{27606}
}

\begin{document}

\begin{abstract}
Hyperparameter tuning is the black art of automatically finding a good combination of control parameters for a data miner. 
While widely applied in empirical Software Engineering, there has not been much discussion on which hyperparameter tuner is best for software analytics.
To address this gap in the literature, this paper applied a range of hyperparameter optimizers (grid search, random search, differential evolution, and Bayesian optimization) to a defect prediction problem.
Surprisingly, no hyperparameter optimizer was observed to be ``best'' and, for one of the two evaluation measures studied here (F-measure), hyperparameter optimization, in 50\% of cases, was no better than using default configurations.

We conclude that hyperparameter optimization is  more nuanced than previously believed. 
While such optimization can certainly lead to large improvements in the performance of classifiers used in software analytics, it remains to be seen which specific optimizers should be applied to a new dataset.

\end{abstract}

 

\maketitle
\section{Introduction}

Recent results from software analytics show that the performance of a data miner exploring software data
is significantly increased after applying hyperparameter tuning \cite{AGRAWAL2018,agrawal17better,Fu2016TuningFS,Fu17easy,Fu16Grid,jia15combinatorial,wang13config,corazza10ee,minky13ensemble,song13ee}. 
Such tuners automatically search the  very large input space of possible  parameters settings for a  data miner.

Researchers in this area use a  very narrow range of optimizers. For example, WEKA comes with a hyperparameter optimizer based on SMAC (a state-of-the-art Bayesian optimization method~\cite{hutter11smac}). While there is much to recommend SMAC, it is only one of a wide range of possible tuners including  grid search,
random search~\cite{Bergstra:2012},    evolutionary modern search methods, sampling methods~\cite{jia15combinatorial},
or various domain-specific methods that exploit some aspect of the local problem~\cite{chen2017riot}.
Before this community uncritically endorses the use of a single tuner, it seems appropriate and timely to reflect on the relative
merits of multiple tuners. Hence, this paper. 

\begin{table*}[!t]
\footnotesize
\centering
\vspace{-0.35cm}
\caption{Object Oriented Measures used in our defect datasets }
\label{tbl:oometrics}
\resizebox{\linewidth}{!}{
\begin{tabular}{|l|l|p{15cm}|}
\hline
\rowcolor{gray!80}Metric & Name & Description \\\hline
amc & average method complexity & Number of JAVA byte codes \\\hline
avg\_cc & average McCabe & Average McCabe's cyclomatic complexity seen in class \\\hline
ca & afferent couplings & How many other classes use the specific class. \\\hline
cam & cohesion amongst classes & Summation of number of different types of method parameters in every method divided by a multiplication of number of different method parameter types in whole class and number of methods. \\\hline
cbo & coupling between objects & Increased when the methods of one class access services of another. \\\hline
ce & efferent couplings & How many other classes is used by the specific class. \\\hline
dam & data access & Ratio of private (protected) attributes to total attributes \\\hline
dit & depth of inheritance tree & It's defined as the maximum length from the node to the root of the tree \\\hline
ic & inheritance coupling & Number of parent classes to which a given class is coupled (includes counts of methods and variables inherited) \\\hline
lcom & lack of cohesion in methods & Number of pairs of methods that do not share a reference to an instance variable. \\\hline
locm3 & another lack of cohesion measure &   $m,a$ =    number of methods,attributes in a class;  $\mu(a)$ =  number of methods accessing an attribute. $lcom3 = ((\frac{1}{a}\sum_{j}^{a}\mu(a_j))-m)/(1-m)$\\\hline
loc & lines of code & Total lines of code in this file or package. \\\hline
max\_cc & Maximum McCabe & maximum McCabe's cyclomatic complexity seen in class \\\hline
mfa & functional abstraction & Number of methods inherited by a class plus number of methods accessible by member methods of the class \\\hline
moa & aggregation & Count of the number of data declarations (class fields) whose types are user defined classes \\\hline
noc & number of children & Number of direct descendants (subclasses) for each class \\\hline
npm & number of public methods & npm metric simply counts all the methods in a class that are declared as public. \\\hline
rfc & response for a class & Number of methods invoked in response to a message to the object. \\\hline
wmc & weighted methods per class & A class with more member functions than its peers is considered to be more complex and therefore more error prone \\\hline
defect & defect & Boolean: where defects found in post-release bug-tracking systems.\\\hline
\end{tabular}}
\vspace{-0.2cm}

\end{table*}

\begin{figure}
\includegraphics[width=3.5in]{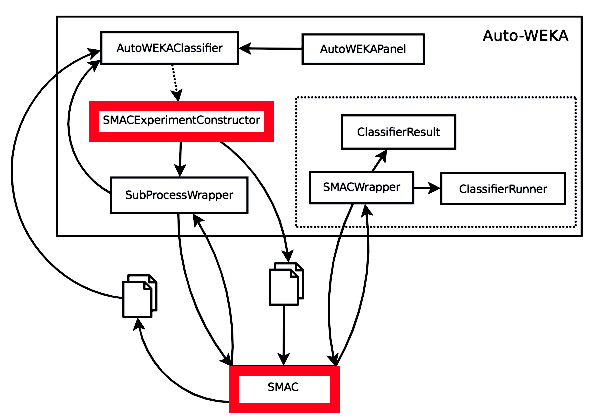}
\caption{WEKA's hyperparemter tuning tool from \href{https://www.cs.ubc.ca/labs/beta/Projects/autoweka/}{UBC}. Note that central to the tool is a single hyperparameter optimizer called SMAC~\cite{hutter11smac},  shown in \textcolor{red}{{\bf RED}}. This paper asks the question ``is one hyperparameter optimizer enough?''. }
\vspace{-10pt}
\end{figure}

The experiments in this paper document the efficacy of default versus tuned settings using four state-of-the-art hyperparameter optimization techniques (grid search, random search, differential evolution, and Bayesian optimization) across four representative classes of data miners (decision tree, random forest, support vector machine, and k nearest neighbors). This study investigate the practicability and benefits of hyperparameter tuning in defect prediction for three goals of (1) F-Measure, (2) Precision\footnote{These
measures were used since they reference multiple target classes and
optimizing for goals based on single targets leads to (e.g.) high
recalls {\em and} false alarms}, and (3) extensively on per software release version level \footnote{Most of the previous studies investigated just on project version \cite{agrawal17better, Fu2016TuningFS}.}. 
Note that we make no claim that the tuners we explore cover the space of all possible hyperparamter optimizers. Instead, we explore just some of the more popular ones and show that there is much variability in which of these tuners is best. This result,
even on just the tuners explored here, is sufficient to motivate future work that explores how to select tuning algorithms
for SE data sets. This work shall explore two  research questions.

\textbf{RQ1:  { Is hyperparameter tuning useful in defect prediction?}}

With current success of state-of-the-art hyperparameter tuning work in software analytics \cite{Fu17easy, Fu2016TuningFS, AGRAWAL2018}, we aim to challenge the versatility of the success with smaller scope of datasets specifically in defect prediction (from project version to software release version). Our experiment shows statistically significant improvements while using hyperparameter tuning in defect prediction (especially in precision). In summary, we show that:       
\vspace{0.25cm}

\begin{result}{ 1}
Hyperparamter tuning is useful in defect prediction. It confirms with the recent success of hyperparameter tuning in empirical SE study.
\end{result}

 This confirmation of the usefulness of hyperparameter tuning practice in defect prediction leads us to wonder which hyperparameter optimizer should be considered as the best standard one. Hence, our next question is as follows.
 
\textbf{RQ2:  { Which hyperparamter optimizer is the best for defect prediction?}}

We find that some optimizers work best only for specific learners and evaluation criteria; e.g. Bayesian Optimization works well in Precision but not in F-Measure while DE optimizes greatly in F-Measure but not in Precision. Moreover, the time cost of Bayesian Optimization is expensive (twice to 100 times longer than other  techniques). Hence we say:
\vspace{0.25cm}

\begin{result}{ 2}
No hyperparamter optimization technique was considered to be best. 
\end{result}

That is, our answer to the question ``is one hyperparemeter optimizer enough'' is ``no''. Hence we
must deprecate papers that report the results of tuning based on a single optimizer.




In summary, the main contributions of this paper are:
\vspace{-0.15cm}

\bi

\item An extensive experimental survey for hyperparameter tuning in defect predictions;
\item A comment on the (lack of) generality
of conclusions from such hyperparameter studies
(we cannot claim that one hyper parameter optimizer is better than another);
\item A reproduction package containing all the data, algorithms, and experimentation of this paper, see \url{https://github.com/ai-se/hyperall}.
\ei

The rest of this paper is organized as follows. Section 2 describes
 background, how defect predictors can be generated by
 data miners, and how tuning can affect the effectiveness of the
 learners. Section 3 defines the experimental setup of this paper. Section 4 presents the results and discussions from
 the case study. Lastly, we discuss the validity of our results 
 and a section describing our conclusions.

\section{Related Work}

\subsection{Defect Prediction}
Human programmers are clever, but flawed. When there are software functionality developments, there must also be software defects. Defects include software crashes or wrong and lack of appropriate functionality. With the inherent existence of defects, it is important to be aware of the defect and take proactive approaches to minimize and prevent future defects. It is imperative to efficiently summarize the knowledge about defects within the system in order to balance with taking proactive action toward defects while developing new functionality to the system. Testing before software is deployed is one approach to learn about the existence of defects within the system. However, according to Lowry et al. \cite{lowry98}, software assessment budgets are finite while assessment effectiveness increases exponentially with assessment effort. According to the 80/20 Pareto principle in software testing, 20\% of the application contains 80\% of the critical defects.
Therefore, in order to preserve the finite resources, the gold standard practice is to apply
the best available resources (labor, knowledge, time, etc) on the critical code portion. Any method that focuses arbitrary parts of the code can
miss critical defects in other areas, which means some sampling policy
should be implemented to smartly explore the rest of the system.

One smart sampling policies class is defect predictors which learned
from static code attributes. Such defect predictors are easy to apply, widely used, and useful. Given object oriented software attributes described
like Table \ref{tbl:oometrics}, data miners can infer where software defects (dependent attribute)
mostly occur and learn the pattern of how defects will occur. Static code attributes can
be automatically collected as independent attributes, even for very large systems \cite{nagappan05}. Otherwise, manual code reviews method can be applied, which is slower and more labor intensive
\cite{menzies02_truism}. Researchers and industrial
practitioners use static attributes to guide software quality
predictions \cite{nam15hetero, tan15online, krishna16bellwether, lewis13_bug}
and such predictors can localize 70\% (or more) of the defects in code \cite{menzies07defect}.

\subsection{Data Miners} 
Defect predictors use data miners to apply various heuristics to efficiently reduce the search space for finding the summaries of the defect data. This study uses 4 popular data miners including Classification And Regression Trees (CART), Random Forests (RF), K Nearest Neighbors (KNN), and Support Vector Machine (SVM).  They are interesting learners in that they represent all the four statistically distinct classes of a performance spectrum for defect predictors that were categorized by Ghotra et al~\cite{ghotra15} through the double Scott-Knott test. Specifically: 

\bi
\vspace{-0.2cm}
\item CART is a tree learner that divide a
data set, then recursively split on each node until some stop
criterion is satisfied. 

\item RF follows the procedure like CART but RF is an ensemble method of building $N$ CART  ($N > 1$), each time using some random subset of the attributes. 

\item KNN is a lazy-learning and instance based technique that studies the $K$ most similar training examples to a particular instance to classify that instance. 

\item SVM uses hyperplane to separate two classes (defective versus non-defective). SVM applies the kernel tricks to make the data more separable which transforms the data points into multi-dimensional feature space.

\ei

This standard of picking four different levels of performance between various data miners was also adapted for other recent empirical SE studies \cite{di18_fft, fu18_fft, krishna17_bell, agrawal17better}.

\subsection{Hyperparameter Tuning}

Data miners have control parameters (e.g., for SVM it would be the kernel function type, the regularization term, the tolerance,
etc.). Adjusting those parameters to optimize the performance of the data miners is called hyperparameter tuning~\cite{Fu2016TuningFS, Fu17easy, AGRAWAL2018, Fu16Grid}. Tuning is used in the hyperparameter
optimization literature exploring better combinatorial
search methods for software testing \cite{jia15combinatorial} or the use of genetic
algorithms to explore 9.3 million different configurations
for clone detection algorithms \cite{wang13config}.
Other researchers explore the effects of parameter tuning on
topic modeling for SE text mining. Tuning is also used for software effort estimation; e.g. using
tabu search for tuning SVM \cite{corazza10ee}; or genetic algorithms for
tuning ensembles \cite{minky13ensemble}; or an exploration tool for quality checking
of parameter settings in effort estimators \cite{song13ee}. 

\pdfoutput=1
\begin{figure}[t]
\vspace{-0.45cm}
\begin{center}
    \includegraphics[width=0.47\textwidth]{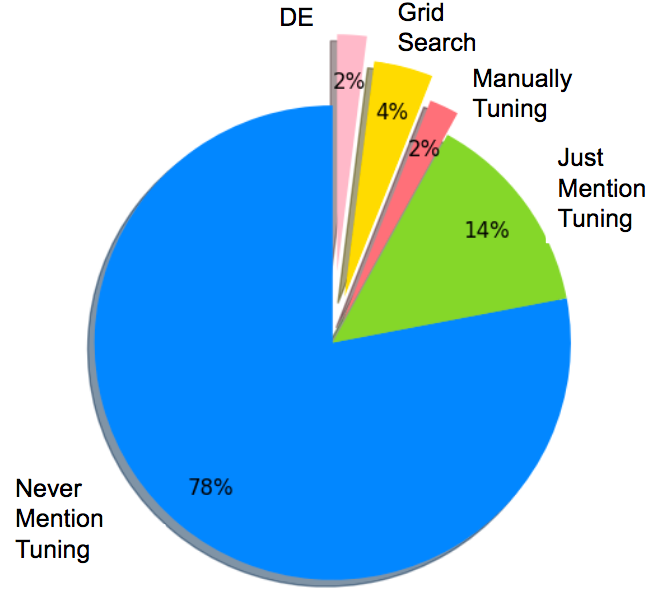}
\end{center}
\caption{Literature review of hyperparameters tuning on 52 top defect prediction papers \cite{Fu16Grid}}\label{fig:litfigure}
\vspace{-0.5cm}

\end{figure}

\begin{table*}[!t]
\centering\footnotesize
\vspace{-0.25cm}
\caption{Parameters Tuning Space}
\vspace{-0.1cm}
\label{tbl:paramspace}
\begin{tabular}{|p{1cm}|c|c|c|l|}
\hline
\rowcolor{gray!80}Learner & Parameter & Default & Tuning Range & Description \\\hline
\multirow{5}{*}{CART} 
& criterion & ``gini'' &  [``gini'', ``entropy''] & The function to measure the quality of a split.\\
& max\_features & None & [0.1, 1.0] & The number of features to consider when looking for the best split. \\
& min\_samples\_split & 2 & [2, 30] & The minimum number of samples required to split an internal node.\\
& min\_samples\_leaf & 1 & [1, 21] & The minimum number of samples required to be at a leaf node.\\
& max\_depth & None & [1, 21] & The maximum depth of the tree.\\\hline

\multirow{2}{*}{KNN}  
  &  n\_neighbors & 5 & [2, 10] & Number of neighbors to use. \\
  &  weights & ``uniform'' & [``uniform'', ``distance''] & Weight function used in prediction. \\\hline
 
\multirow{4}{*}{SVM}  
 & C & 1.0 & [1, 100] & Penalty parameter C of the error term. \\
 & kernel & ``rbf'' & [``rbf'', ``sigmoid''] & Kernel type to be used in the algorithm. \\
 & coef0 & 0.0 & [0.1, 1.0] & Independent term in kernel function.   \\
 & gamma & 'auto' & [0.1, 1.0] & Kernel coefficient.  \\\hline

\multirow{5}{*}{RF}  
 & criterion & ``entropy'' &  [``gini'', ``entropy''] & The function to measure the quality of a split.\\
& max\_features & 'auto' & [0.1, 1.0] & The number of features to consider when looking for the best split.\\
& min\_samples\_split & 2 & [2, 30] & The minimum number of samples required to split an internal node.\\
& min\_samples\_leaf & 1 & [1, 21] & The minimum number of samples required to be at a leaf node.\\
& n\_estimators & 10 & [10, 100] & The number of trees in the forest.\\\hline
 
\end{tabular}
\end{table*}

The case studies used in this paper comes from defect predictor or classification of existing static code attributes. Many SE defect prediction studies on static code attributes have been produced \cite{krishna16bellwether, nam15hetero, tan15online}. However, software analytic practitioners have only been solely focusing on finding and employing complex and ``off-the-shelf'' machine learning models \cite{menzies07defect, moser08defect, elish08defect}. According to literature reviews done by Fu et al \cite{Fu16Grid} in defect prediction shown in Figure \ref{fig:litfigure}, 80\% of highly cited papers did not mention any parameters tuning while employing the default parameters setting of the data miners. 

Bergstra and Bengio \cite{Bergstra:2012} noted on the popularity of grid
search: (a) simple search to give some degree
of insight; (b) has little technical overhead; (c) simple to automate and parallize; (d) (on a computing
cluster) can find better tunings than sequential optimization. Grid search is conjectured not more effective than more randomized searchers if the underlying search space dimension is inherently low.

Tantithamthavorn et al. \cite{tanti16defect} and Fu et al. \cite{Fu2016TuningFS} are two recent work investigating the effects of parameter
tuning on defect prediction, Tantithamthavorn used grid search while Fu
applied differential evolution.  Neither offer a comparison
of their preferred tuning method to any other. At the same time, Fu et al (1) only studied half of the 4 data miners classes Ghotra et al considered; (2) did not include the state-of-the-art hyperparameter tuning Bayesian Optimization method; and (3) applied the optimization on project level instead of the release version level. 

Beside the strength or circumstance to pick the right optimizer, tuning for many objectives or inappropriate goals at one dispersed the strength of optimization. From Sayyad et al's results through tuning multi-objectives in effort estimation, all the methods did similarly in tuning 2 objectives but most of these algorithms do not perform nearly acceptably in tuning 4-5 objectives \cite{sayyad13sbse} when comparing the percentage of fully-correct solutions in the Pareto fronts. However, Recent study by Agrawal et al \cite{agrawal17better} determined that better data quality is more important than better data miners quality by tuning the data preprocessors. It is apparent because hyperparameter tuning can be applied for not only the data mining but also the preprocessing data (SMOTE, SMOTUNED \cite{AGRAWAL2018}, normalization, discretization,
outlier removal, etc) and features selection (explore $2^N$ subsets of $N$ features with PCA, RFE, etc). Thus, the appropriate goals of tuning or/and knowing what to tune are important.

The lack of these points in SBSE's literatures basically stemmed from Lessmann et al's conclusion \cite{lessmann08benchmarking} as software analytics practitioners are flexible to pick from a broad set of models when building defect predictors since the importance of the data miner is generally not too significant. Knowing that, is the insignificant difference in performance due to the nature of defect prediction problem itself or the traditional approach of exploring the tuning input space to the problem? For instance, Fu, Chen, and Agrawal \cite{fu18_fft, di18_fft, agrawal18_fft} had applied simple method designed by psychological principles, Fast and Frugal Trees \cite{phillips17_fft}, that focusing on exploring the output space as binary tree with depth $d = 4$ instead of exploring the input space. This backward approach offers better or similar performance (for most cases) but with much less trade-off in term of time, processing power, and result's human-readability. Consequently, it is important to assess the old belief, which has pushed for this more extensive guideline and survey of hyperparameter tuning in defect prediction.

\section{Experimentation}

For each tuning goal (precision and F1), each tuning algorithm shall run 20 times across the four machine learner models (CART, KNN, SVM, and RF) to validate the stability of the results across through random biases and noises. For each repeat of the algorithm, different combination of hyperparameter settings would be evaluated. Each evaluation is quantified as 10 parameters sets generated by corresponding tuning algorithm. Each evaluation  would be compared against the current ``best'' one. If better, then it will replace the ``best'' one. If not, it would be less likely that the next evaluation would achieve higher ``best'' so the lives of the process is decreased by 1. The stop conditions include the exhaustion of five lives or 1 hour processing time, the search process is repeated till either the stop condition meets. For each release version $i$ of project $j$, the results from training the respective scoring goal with data miner $k$ with all tuning methods on release version $i+1$ would be recorded and ranked by the Scott-Knott test. The tuning method that statistically got first rank will be incremented by one to measure how often each optimizer would statistically perform best.  

\subsection{Tuning Algorithms} 

This study shall explore the representatives of hyperparameter tuning classes including: grid search, random search, DE, and Bayesian optimization. All of these optimization methods explored the parameter space as
described in Table~\ref{tbl:paramspace}.

{\em  Grid Search} is simply picking a set of values for each configuration
parameter and evaluating all the combinations of these
values, and then return the best one as the final optimal result,
which can be simply implemented by nested for-loops. 

{\em  Random search} is nothing but randomly generated set of candidate parameters from the same tuning range as in Table \ref{tbl:paramspace}. 

{\em  DE} evolves a new generation of candidates by extrapolating randomly between three current population's members of solutions, $\mathit{pop}_i$, of size $\mathit{np}$~\cite{storn1997de}. DE combines local search mutation, \mbox{$y_k = a_k + f \times (b_k - c_k)$} (where $f$ is a parameter controlling crossover), with an archive pruning operator. As the process progresses, new candidates $y$ supplant older items in the population, then all subsequent mutations use the newer and more valuable candidates. 


{\em  Bayesian Optimization} comprises a probabilistic model and an acquisition function. There are several popular approaches for probabilistic models: density estimation models such as Tree-structured Parzen Estimator (TPE) \cite{bergstra11TPE}, random forest such as Sequential Model-based Algorithm Configuration (SMAC) \cite{hutter11smac}, and Gaussian process \cite{snoek12gp}. Specifically, this study employed random forest based probabilistic method,  SMAC \cite{hutter11smac}. It is known as auto-sklearn, a robust AutoML system based on scikit-learn \cite{scikit-learn}, developed by Feurer et al \cite{feurer15autosk} which discards poorly performing hyperparameter early. A new input's posterior mean and variance of is computed then used for computation of the acquisition function. The acquisition function defines the criterion to determine future parameters candidates for evaluation. The next most promising parameters set will be found using the probabilistic model, evaluated by acquisition function, updated within the main model, and reiterated. 

\begin{figure*}[!t] 
        \centering
        \includegraphics[width=.8\linewidth]{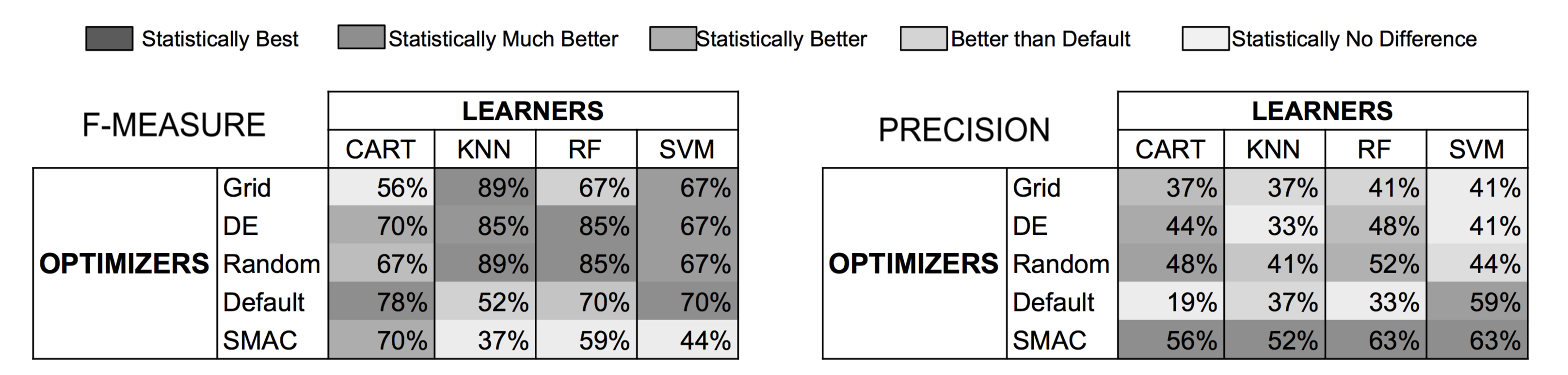}
    \caption{Comparison of applying different methods of tuning against default setting of various data miners on 27 release versions of 10 projects. The colors refer to a statistical comparison across the tuning performance of optimizers in each row of learners.}
    \label{fig:results}
\end{figure*}

\subsection{Defect Data} Our data comes from SEACRAFT repository (tiny.cc/seacraft).
This data pertains to open source JAVA systems defined in terms: \textit{ant}, \textit{camel}, \textit{ivy}, \textit{jedit}, \textit{log4j}, \textit{lucene}, \textit{poi}, \textit{synapse}, \textit{velocity} and \textit{xerces}. 



We applied incremental learning approach. With at least three software
releases (where release i+1 was built after release i), this will allow defect predictors being built to predict future (test) defects based on learning from the past (train) data. Specifically: 

\bi

\item Each software release $i$ was divided into 3 even portions ($i/3$) where a learner will be trained on $2/3$ of $i$ and each candidate of parameter setting would be evaluated on the other $1/3$ of $i$. 

\item After terminating the tuning method, the best parameters setting shall be picked for building the appropriate data miner model on the release $i$ to predict the defects in release $i+1$ and output the result according to the tuning goal. 

\item These 4 data miners will also be trained with also default parameters configuration on release $i$ then tested on release $i+1$.

\ei

\subsection{Tuning Goals}  

The problem studied in this paper is a binary classification task of bug identification based on the static code attributes of a specific JAVA class. The performance of a binary classifier can be assessed via a  confusion matrix as in Table~\ref{fig:cmatrix}.

\begin{wraptable}{r}{1.5in}
\small
\begin{center}
\vspace{-0.45cm}
\caption{Confusion Matrix} 
\vspace{-0.1cm}
\label{fig:cmatrix}
\begin{tabular} {@{}cc|c|c|l@{}}
\cline{3-4}
& & \multicolumn{2}{ c| }{Actual} \\ \cline{2-4}
& \multicolumn{1}{ |c| }{Prediction} & false & true  \\ \cline{2-4}
& \multicolumn{1}{ |c| }{negative} & $\mathit{TN}$ & $\mathit{FN}$ & \\ \cline{2-4}
 & \multicolumn{1}{ |c| }{positive} & $\mathit{FP}$& $\mathit{TP}$  &  \\ \cline{2-4}
\cline{2-4}
\end{tabular}
\end{center} 
\end{wraptable}

Further, ``false'' means the learner got it wrong and ``true'' means the learner correctly identified
a positive or negative class. Hence, Table~\ref{fig:cmatrix} has four quadrants containing, e.g., $\mathit{TP}$ which denotes ``true positive''.

Our optimizers explore tuning improvements for {\em Precision} and {\em F-Measures} values on software release version level. For these two goals, the larger the values, the better the model's predicting power.    

\vspace{-5pt}
\begin{equation}
 \mathit{Precision} = \frac{\mathit{TruePositives}}{\mathit{TruePositives} + \mathit{FalsePositives}}
 \end{equation}
 \vspace{-10pt}

 \begin{equation}
 \mathit{F-Measure} = \frac{2 * (\mathit{Recall} * \mathit{Precision})}{\mathit{Recall} + \mathit{Precision}}
\end{equation}

We do not explore all goals since some have trivial, but not
insightful, solutions. No  evaluation  criteria  is  ``best''  since  different  criteria  are appropriate  in  different  real-world  contexts. For example, when we tune for recall, we can
achieve near 100\% recall but at the cost of a near 100\% false
alarms.  Precision's definition takes into accounts not only the
defective examples but also the none defective ones as well so
it has this effect of where multiple goals are in contention. The same is true for the F-Measure
(as it uses precision).



\subsection{Statistical Analysis}
We compared our results of tuned miners per release version using statistical significance test and an effect size test by Scott-Knott procedure \cite{mittas2013ranking, ghotra15}. Significance test detects if two populations differ merely by random noises \cite{ghotra15}. Effect sizes checks whether two populations differ by more than just a trivial amount, where $\mathit{A12}$ effect size test was used \cite{arcuri2011practical}. Our stats test are statistically significant with 95\% confidence and not a ``small'' effect ($\mathit{A12} \ge 0.6$).




\section{Results}

Figure \ref{fig:results} offers the cumulative statistical results of Grid, DE, Random, and SMAC against the default hyperparameter configuration for each data miner across the 27 release versions to maximize \textit{Precision} and \textit{F-measure} scores for this study. There are 27 release versions which correspond to 27 possible times that one tuning method can get first rank. For example, CART tuned with SMAC got first rank 19 times (70\%) while CART tuned with Grid search only got first rank 11 times (56\%). The \textit{darker} the cell, the statistically \textit{better} the performance of the learner combined with that optimizer. 

From Figure \ref{fig:results}, we observe that over all 
{\em (learners, optimizers,  evaluation criteria)}, there is no clear ``best'' optimizer:
\be
\item Grid search, widely depreciated~\cite{Bergstra:2012}, performs surprisingly well for KNN and SVM's {\em F-Measure} (but not elsewhere);
\item
 DE does well
for optimizing {\em F-Measure} but not {\em Precision};
\item
 Bayesian Optimization, SMAC, gets best results in {\em Precision}, but not for   {\em F-Measure}; 
 \item
 Other optimizers work best only for specific learners and evaluation criteria.
\ee

Further to the third point, Table~\ref{tbl:runtimes}
shows the mean CPU time in seconds to run one repeat
of one optimizer on one learner. Note that the SMAC
runtimes are substantially larger than the other methods.
 Hence the extra benefits of SMAC optimization
must be carefully 
weighed against the cost of that optimization. \pdfoutput=1
\begin{table}[!t]

\caption{Runtime in seconds.}\label{fig:runtime}
\vspace{-5pt}


\footnotesize
\begin{tabular}{r|cccc}


 & ~CART & ~KNN & ~RF & ~SVM \\ 
\hline
Grid    &  4  &   5    & 334 & 6 \\ 
DE      & 4  &   5    & 318 & 6 \\  
Random  & 4  &   5    & 305 & 6 \\ 
Default & 1  &   1    & 2   & 1 \\  
SMAC    & 613     &   501    & 652 & 505  \\

\end{tabular}
\vspace{-17pt}
\label{tbl:runtimes}
\end{table}
 For example,
for {\em Precision} and SVM, is a win of 63\% (with SMAC) vs 59\% (with Defaults) really worth the CPU required
to earn such a small gain?

If the reader feels that the CPU times recorded
in Table \ref{tbl:runtimes} are insignificantly small,
then please recall that these are  for $R=1$ repeats over  $L=1$ learners for $O=1$ optimizer for $D=1$ datasets.
When repeated for larger $R \times L \times O \times D$ values,
the longer runtimes of SMAC become highly significant. 
For example, even
utilizing our university's cloud compute facilities, 
it took two graduate students weeks to collect the data behind
Figures~2 \& 3.

\section{Discussion}

\subsection{Algorithm}

Even when there seems to have no conclusive evidence to indicate which optimizer is the best across the three goals. DE did well to optimize F-Measure while SMAC did well to optimize Precision on per the release version level. Naturally, a thorough investigation of all options via grid search should do better than a partial exploration of just a few options, through DEs and SMACs. 

In reality, both grid search and random search sample through different parameter settings
between some $min$ and $max$ value of predefined tuning range, which will determine the nature of tuning and good tuning require expert knowledge. If the best options lie in between these jumps, then grid
search will skip the critical tuning values. Moreover, for both grid search and random search, all the combination options in the predefined tuning range are independently
evaluated. Any lessons learned in the process will not be utilized to improve in the remaining runs.

Note that both DE and SMAC are more prone to not skip and incrementally fill in the gaps between initial selected tuning range. Moreover, both evolutionary nature of DE and Bayesian nature of SMAC, learning knowledge are transferred to the next generation in the same run to improve the inference of future results: 

\bi
\item For DE, tuning values are adjusted by some random amount that
is the difference between two randomly selected vectors. DE's discoveries
of better vectors accumulate in the frontier, new solutions (candidates) are being continually built
from increasingly better solutions cached in the frontier.

\item For SMAC, given a
small initial set of function evaluations, proceeds by fitting a surrogate model to those observations,
random forest like (SMAC), and then optimizing
an acquisition function that balances exploration and exploitation. With randomness and probability distributions, it determines the next most promising point to evaluate. 

\ei

Our results aligned with Bergstra and Benigo's formal analysis for how
random searches (like DEs and SMAC) can do better than grid search and random search especially if the
region containing the useful tunings is very small. In such search space: (a) Grid search and Random search can waste much time exploring
an irrelevant part of the space. (b) Grid search's effectiveness is
limited by the curse of dimensionality. 

\subsection{Approach}

The core experiment can be seen as narrow - by looking solely at defect prediction - but is indeed appropriate and necessary to start a discussion on the complexity and potential limitations of parameter tuning methods. Introducing hyperparameter tuning can come with a great trade-off of complexity and cost (processing power and time) if the goals are not achieved within a reasonable cost. More important, hyperparameter tuning can be incorporated beside the data mining step such as during the preprocessing data and features selection. Moreover, Table \ref{tbl:paramspace}'s tuning space can be expressed  continuous which means
the space of parameters is theoretically infinite. It is reasonable that the complexity is unwanted unless the area that needed to be tuned is known. 

In Calero's and Pattini's survey of modern SE companies \cite{calero15_green}, they
find that many current organizational redesigns are motivated (at
least in part) by arguments based on ``sustainability'' (i.e., using
fewer resources to achieve results). According to them, ``redesign for greater simplicity'' is a new source of innovation and motivation for much contemporary industrial work to explore cost-cutting opportunities for gaining an advantage over other competitors. 
Perhaps, it is time to call for a new approach to software analytic beside the traditional \textit{forward} approach of exploring the input space that we followed for this study.

\section{Threats to Validity}
Biases are inevitable in any empirical study that can affect the results. Although, this work has attempted to minimize biases, the following issues should be considered when inferring insight from the results. 

\subsection{Order Bias}

With  each  dataset  how  data  samples  are  distributed  in  training  and  testing  set  is  completely  random.Though  there  could  be  times  when  all  good  samples  arebinned  into  training  set.  The experiment is designed to run for 20 times in order to mitigate that bias and for stability of the results. For each run, the
random seed is different for each data set, but it
will be the same across learners configured by hyperparameter optimizers for the same data set.  With this approach, it is important to note that different triplets have different seed values (so this case study does sample across a range of search biases). 

\subsection{Sampling Bias}

Sampling Bias threatens any classification experiment, i.e.,what proves to be important here may be insignificant there. For e.g., We applied ten widely used
open source JAVA software project data from SEACRAFT as
the subject in various  case  studies  by  various  researchers \cite{Fu2016TuningFS, di18_fft, fu18_fft}, i.e.,  our  results  are  not  more  biased  than  many  other  studies in this arena. However, the datasets were supplied by one individual. Moreover, only specified metrics listed in Table 2 are used as the attributes to build defect predictors, it is not scientific to guarantee that
our observation can be directly generalized to other projects that
using different set of metrics, like code change metrics. 

Also, these defect datasets are lower in dimension in comparison with effort estimation, text mining, and test case prioritization. Consequently, our findings for this specific case study in defect prediction might not be to applicable in those other software analytics.

\subsection{Learner and Optimizer Bias}

Research reproducability refers to the consistency of the results reproduced and obtained
from this particular designed experiment. To assure the research reproducibility and reliability,
this paper has taken care to either clearly define our algorithms
or use implementations from the public domain (Scikit-Learn). While most of our optimizers were solely developed based on the public domain foundation, there are different algorithm based implementation algorithms with Bayesian Optimization (SMAC) in Auto-Sklearn and DE which may affect results differently if other algorithm implementations were considered, i.e. for  probabilistic models of Bayesian Optimization, TPE or Gaussian Process can be incorporated instead. However, in term of runtime cost, the data loading and processing methods implemented in this study are used by all optimizers. Therefore, the relative runtime cost comparison between all still hold.

\subsection{Evaluation Bias}

The SMAC optimizer's internal configuration, auto-sklearn, forestalls the incorporation of many non-conventional evaluation metrics \cite{feurer15autosk}. This work was only able to report on two performance measures, Precision and F-Measure (as defined in equation 1 and 2), on per software version level. Other quality measures (AUC, $P_{opt}$, Distance2Heaven, etc) are also often used
in software engineering to quantify the effectiveness of prediction \cite{ di18_fft, monden13_assessing, kamei13_assurance}. 

\section{Conclusion and Future Work}

Three conclusions following from this extensive hyperparameter tuning study in defect prediction.
Firstly,   
like many other researchers before us~\cite{Fu2016TuningFS, Fu16Grid,jia15combinatorial,wang13config,corazza10ee,minky13ensemble,song13ee,AGRAWAL2018, agrawal17better, Fu17easy},
we conclude that 
{\em hyperparameter optimization is very useful}.
Figure \ref{fig:results} is very clear: hyperparamter tuning
usually leads to far better performance scores
than just using  the defaults.

That said, our  second conclusion is that 
{\em it is not clear when one  
hyperparameter optimizer is better than any other}.
Hence, for future researches, practitioners would need to apply a range of optimizers
rather than rely on just one. It is similar to Lessmann et al's conclusion \cite{lessmann08benchmarking} as software analytics practitioners are flexible to pick from a broad set of models when building defect predictors since the importance of the data miner is generally not too significant.

So thirdly,
{\em reducing the total runtimes of 
multi-optimizer studies is an open and pressing problem}.
We cannot expect a wide community of academic
and industrial practitioners to use hyperparameter optimization {\em unless} that usage is easy to apply.



As for future work, we propose:




\bi
\item This study can be replicated with other evaluation measures, i.e. AUC, $P_{opt}$, Distance2Heaven, etc. Bayesian Optimization practice (SMAC) of auto-sklearn can be edited to adapt those evaluation measures while optimizing the data miners.  
\item This work should be repeated for different domains in software analytics such as text mining, effort estimation, etc.
\item Other parts of the data mining process can be explored for optimization (preprocessing, features engineering, etc)
\item Research to determine how to check if a specific problem is tunable? And if so, which part of the data mining pipeline along which goals should be tuned?
\item Explore more of the ``backward approach'' of surveying the result space by some initial and random data mining, then reflecting and redesigning a software quality predictor that better understands the results space.

 \ei

\balance

\bibliographystyle{ACM-Reference-Format}


\bibliography{main.bbl}

\end{document}